\documentclass{article}

\usepackage[preprint]{neurips_2022}

\usepackage[utf8]{inputenc} 
\usepackage[T1]{fontenc}    
\usepackage{hyperref}       
\usepackage{url}            
\usepackage{booktabs}       
\usepackage{amsfonts}       
\usepackage{amsmath}
\usepackage{nicefrac}       
\usepackage{microtype}      
\usepackage{xcolor}         
\usepackage{algorithm}
\usepackage{algpseudocode}
\usepackage{wrapfig,booktabs}
\usepackage{subcaption}
\usepackage{graphicx}
\usepackage{ulem}
\usepackage{color, colortbl}

\setcitestyle{numbers}

\title{FedILC: Weighted Geometric Mean and Invariant Gradient Covariance for Federated Learning on Non-IID Data}

\author{Mike He Zhu \\
  McGill University \\
  Mila - Quebec AI Institute\\
  \texttt{he.zhu@mila.quebec} \\
  \And
  Léna Néhale Ezzine \\
  Mila - Quebec AI Institute\\
  \texttt{lena-nehale.ezzine@mila.quebec} \\
  \And
  Dianbo Liu \\
  Mila - Quebec AI Institute\\
  \texttt{dianbo.liu@mila.quebec} \\
  \And
  Yoshua Bengio \\
  Université de Montréal \\
  Mila - Quebec AI Institute\\
  \texttt{yoshua.bengio@mila.quebec} \\
}

\begin{document}

\maketitle
\begin{abstract}
Federated learning is a distributed machine learning approach which enables a shared server model to learn by aggregating the locally-computed parameter updates with the training data from spatially-distributed client silos. Though successfully possessing advantages in both scale and privacy, federated learning is hurt by domain shift problems, where the learning models are unable to generalize to unseen domains whose data distribution is non-i.i.d. with respect to the training domains. In this study, we propose the Federated Invariant Learning Consistency (FedILC) approach, which leverages the gradient covariance and the geometric mean of Hessians to capture both inter-silo and intra-silo consistencies of environments and unravel the domain shift problems in federated networks. The benchmark and real-world dataset experiments bring evidence that our proposed algorithm outperforms conventional baselines and similar federated learning algorithms. This is relevant to various fields such as medical healthcare, computer vision, and the Internet of Things (IoT). The code is released at \url{https://github.com/mikemikezhu/FedILC}.
\end{abstract}

\section{Introduction}

Federated learning is a machine learning paradigm where the local clients collaboratively train a model under the orchestration of a central server  \cite{mcmahan2017communication,jakub2015federated,jakub2016federated}. Keeping the training data spatially-distributed, federated learning remarkably mitigates the privacy risks engendered by the centralized machine learning approaches \cite{mcmahan2017communication,carlini2018secret,duchi2012privacy,dwork2014foundations}. Stupendous achievements though it has obtained, federated learning has been suffocated with the domain shift problems, which occurs when the labelled data collected by the source node statistically differs from the target node's unlabelled data, thus the learning models are unable to generalize to unseen domains with inconsistent data distributions \cite{li2019federated}. In a real-world scenario, for example, although the medical radiology images collected in different hospitals have uniformly distributed labels, the image appearance (e.g. contrast and saturation) may vary due to different imaging machines and protocols used in hospitals. Consequently, the disease detection model collaboratively trained in these hospitals might be unable to effectively generalize to other hospitals with heterogeneity in the feature distributions.

Recently, a slew of researchers endeavoured to devise a methodology to disentangle the domain shift problems by training an invariant model that could successfully adapt to out of domain distributions. The topic of research, also known as "Domain Generalization" or "Out-of-Distribution (OOD) Generalization", aims at lubricating the process of learning invariant features that are generalizable out of domains \cite{candela2009dataset,david2010domain,koby2008learning,kouw2018domain,daume2009easy}. Further, most recent works promote invariance via a regularization criterion at either representation level or gradient level. The former enforces the agreement by generating latent variables that efficaciously represent all the training domains \cite{arjovsky2020irm,bai2020decaug,pezeshki2021gradient,koyama2020ood}, whereas the latter promotes the agreement among the gradients from different domains \cite{parascandolo2021ilc,shi2021gradient,shahtalebi2021sand,rame2021fishr}.

Since the conventional federated learning algorithm involves averaging the model parameters or gradient updates to approximate the optimal global model, our research focuses on facilitating the invariance of gradients among different federated clients. Recently, Parascandolo unearthed that the arithmetic averaging of Hessians is unable to capture the consistencies, thus proposing the geometric mean to encourage the agreements of environments. As opposed to arithmetic mean that performs a "logical OR" on the Hessians, geometric mean serves as a "logical AND" operator, mandating complete consistency among different training domains \cite{parascandolo2021ilc}. Albeit the geometric mean of gradients theoretically promotes consistency, it is still circumscribed with several quandaries. One conspicuous predicament is that the geometric mean becomes legitimate only when all the signs are consistent - either all gradients are positive or negative. Thus, it is unattainable to calculate the geometric mean with \textbf{inconsistent signs}, which commonly exist in non-convex settings. To forestall this issue, Parascandolo, based on his algorithm implementation \footnote{The code implementation of Learning explanations that are hard to vary: \url{https://github.com/gibipara92/learning-explanations-hard-to-vary/blob/main/and_mask/and_mask_utils.py}}, takes \textbf{the absolute values} of all the gradients to calculate the geometric mean, which largely overlooks the inconsistency of signs for the gradients, thus engendering bias during the training process. Meanwhile, Parascandolo also presented a new strategy, namely "AND-Mask", which takes a discrete mask to zero out the gradients with inconsistent signs across environments and ensures the network parameters are updated only if all the gradients from different domains agree on a certain direction. This masking strategy is analogous to applying a logical AND operator on the direction of gradients. Though proved advantageous in some curated test conditions, their proposed mechanism is intrinsically using arithmetic averaging of Hessians. Furthermore, AND-masking on the average loss landscape might induce "dead zones", where the gradients from different environments are sometimes unable to update the parameters unless their directions strictly match with each other \cite{shahtalebi2021sand}. 

Likewise, Rame \textit{et al.} proposed Fishr to approximate the domain-level gradient covariance, which forces the model to have similar domain-level Hessians, thus promoting invariance across domains \cite{rame2021fishr}. However, Fishr could be flawed, especially in a centralized way of training. It requires knowing the number of environments beforehand to calculate gradient covariance, which is unfeasible in an intra-silo training environment. Fortunately, the number of domains is accessible in federated networks, allowing Fishr to capture the inter-silo consistencies in federated learning settings. Shoham et al. proposed a federated curvature (FedCurv) algorithm. Similar to the Fishr method, while applying in a decentralized way, FedCurv employs the diagonal of the Fisher information matrix to mitigate the domain shift problem in federated learning \cite{shoham2019overcome}.

Inspired by geometric mean, Fishr, and FedCurv algorithms, we introduce a new regularization, the Federated Invariant Learning Consistency (FedILC), which leverages the domain-level gradient covariance and the geometric averaging of Hessians to promote both inter-silo and intra-silo consistencies in federated networks. Moreover, we also propose a novel weighted geometric mean mechanism to calculate geometric mean with inconsistent signs. The benchmark and real-world dataset experiments manifest that our proposed algorithm outperforms both the conventional and the state-of-the-art federated learning algorithms.

In a nutshell, the paper purveys the following contributions:

\begin{itemize}

\item We introduce a novel regularization methodology, FedILC, employing the gradient covariance and weighted geometric mean of gradients to capture both inter-silo and intra-silo consistencies in federated networks.

\item We propose the weighted geometric mean to unravel the issue of geometric mean calculation with inconsistent signs on gradients.

\item We proffer a scalable yet straightforward implementation of our algorithm, and the experiments evince that FedILC outperforms both the conventional baselines and other similar federated learning algorithms.
\end{itemize}

\section{Related Work}

\subsection{Strategies of Domain Adaptation}

\textbf{(1) Representation-level agreement:} This approach strives for finding invariant representations at feature level \cite{arjovsky2020irm,bai2020decaug,pezeshki2021gradient,koyama2020ood} by yielding consonant representations for all training domains, thus training models to map different domains into a single statistical distributions. This could be achieved by matching the mean and variance of representations across different domains \cite{sun2016deep}, or the distribution of the representations in different domains \cite{li2018domain}.

\textbf{(2) Gradient-level agreement:} Another approach, on the contrary, works at gradient level, which intends to find local or global minimums commonly shared across all the environments in the loss space \cite{parascandolo2021ilc,shahtalebi2021sand,shi2021gradient}. In other words, the goal minimums are agreements among the gradients backpropagated in the network, and allow it to share similar Hessians for different domains. Specifically, Parascandolo \textit{el al.} proposed geometric averaging of Hessians to capture the invariance among different domains \cite{parascandolo2021ilc}. Compared with arithmetic mean, which operates "logical OR" on the Hessians, geometric mean performs a "logical AND" operation, and achieves better outcomes in finding consistencies among environments. Nonetheless, the geometric mean is intrinsically blemished when the signs are inconsistent, which can be prevalent in non-convex settings. Furthermore, he also applied an AND-masking strategy to the arithmetic mean, which defines a peculiar agreement threshold as a hyper-parameter, and allows the parameters of network to be updated only if all the gradients flowing in that parameter agree on a certain direction based on the threshold. Nevertheless, the masking strategy requires the direction to which the gradients from different environments point strictly match with each other in order to allow the gradients to update that particular parameter. To address this issue, Soroosh proposed a smoothed AND-masking algorithm (SAND-mask), which not only validates the invariance in the direction of gradients, but also promotes the agreement among the gradient magnitudes \cite{shahtalebi2021sand}. Yet, regarding their experiment results, the proposed masking strategy barely competes the performance of other algorithms, including AND-masking strategy. Moreover, Rame proposed Fishr in his paper to leverage the domain-level gradient covariance to promote invariances across domains \cite{rame2021fishr}. However, it is paramount to know the number of environments beforehand to calculate gradient covariance across domains, which is intractable in the centralized way of training.

\subsection{Domain Shift in Federated Learning}

Federated learning aims to train a global model by aggregating the updates from a slew of client models spatially distributed on different devices while preserving privacy. In 2017, McMahan et al., for the first time, presented the notion of federated learning based on data parallelism and proposed both FedSGD and FedAvg algorithms. \cite{mcmahan2017communication}. FedSGD algorithm updates the global model by averaging the gradients uploaded by the clients, whilst the FedAvg methodology employs the averages of clients' model parameters to update the global model's weights. While federated learning provides a nascent methodology in privacy preservation, many challenges arise in terms of the domain shift problem engendered by the non-i.i.d data. Although the authors in \cite{mcmahan2017communication} contend that FedAvg can mitigate the domain adaptation issue to a certain degree, myriads of research have evinced that a deterioration in the federated model's accuracy is almost inevitable on the heterogeneous data \cite{zhao2018fl}. The state-of-the-art approaches to alleviate the non-i.i.d. problems in federated learning can be categorized into the data-based \cite{tuor2020overcoming,yoshida2020hybrid,duan2019self,shin2020xor}, algorithm-based \cite{alireza2020personalized,chen2019meta,hanzely2021fl,dinh2020moreau,arivazhagan2019personalization,shoham2019overcome}, and system-based approaches \cite{mansour2020three,kopparapu2020sequential,ghosh2020robust,ghosh2020efficient}. Precisely, among the algorithm-based methodology, Shoham et al. proposed a federated curvature (FedCurv) algorithm, which is similar to the Fishr paper, while in a decentralized way to disentangle the non-i.i.d issue in federated learning \cite{shoham2019overcome}. During each round, participants transmit updated models together with the diagonal of the Fisher information matrix, which represents the most informative features for the current task. A penalty term is added to the loss function for each participant to promote the convergence towards a globally shared optimum.

\section{Methodology}

\begin{figure}[!htb]
  \centering
  {\setlength{\fboxsep}{0pt}%
  \setlength{\fboxrule}{0pt}%
  \fbox{\includegraphics[width=15.0cm, height=5.0cm]{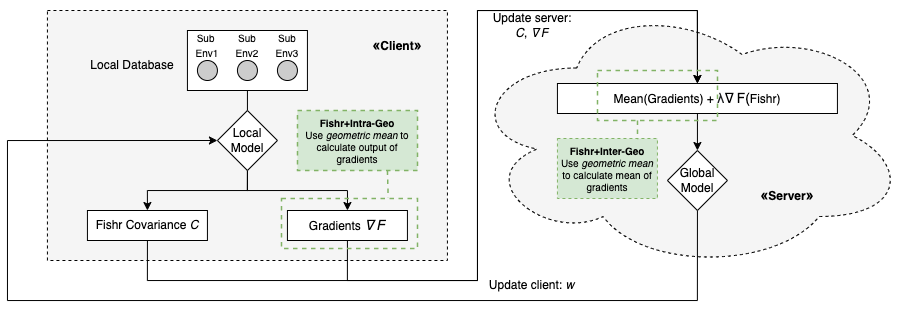}}
}%
  \caption{Server-client communication of Fishr+Inter-Geo and Fishr+Intra-Geo in the federated learning settings.}
 \label{fedilc}
\end{figure}

\begin{figure}[!htb]
  \centering
  {\setlength{\fboxsep}{0pt}%
  \setlength{\fboxrule}{0pt}%
  \fbox{\includegraphics[width=12.5cm, height=20cm]{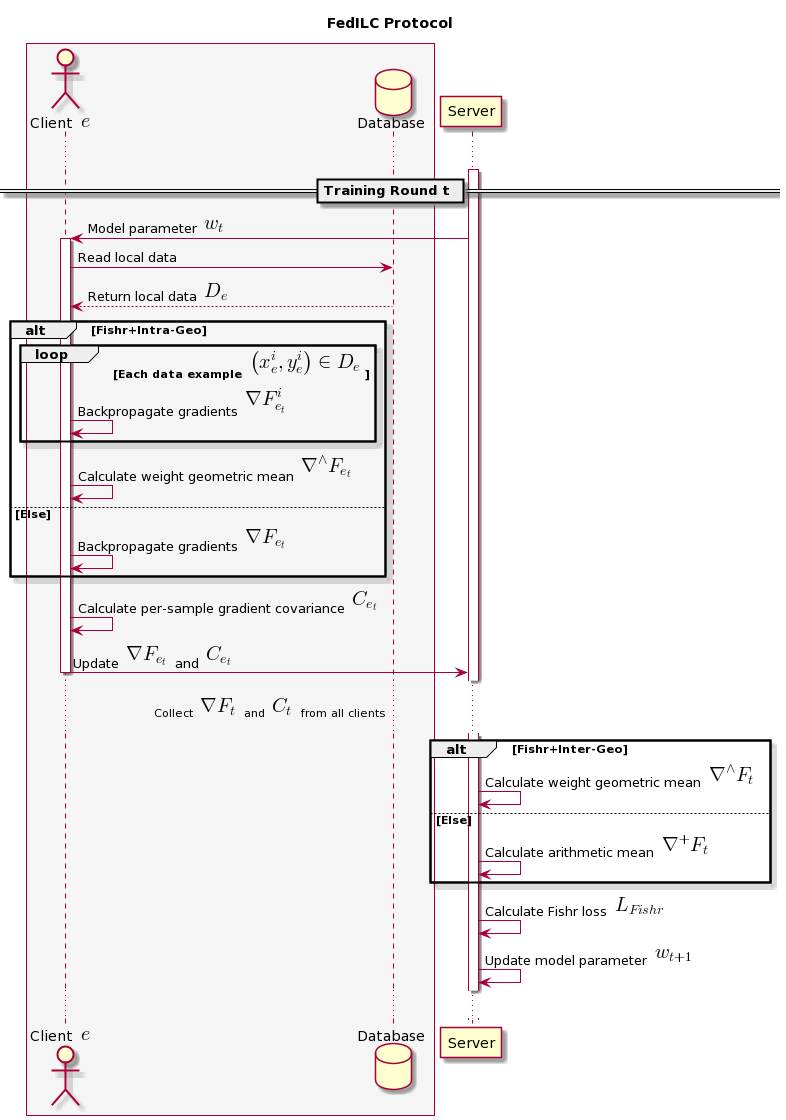}}
}%
  \caption{Sequence diagram to calculate either Fishr+Inter-Geo or Fishr+Intra-Geo algorithm.}
 \label{sequence}
\end{figure}

The domain shift problems engendered by the non-i.i.d. data significantly hinder the training of federated models. Moreover, although the dataset within each client is theoretically homogeneous in distributions, in the real world cases, however, non-i.i.d. data may still exist even within one federated client. Motivated by ILC, Fishr, and FedCurv, we propose two innovative methodologies: (1) Fishr+Inter-Geo, and (2) Fishr+Intra-Geo, to mitigate the inter-silo and intra-silo domain shift problems. Furthermore, we also introduce a novel weighted geometric mean method to calculate the geometric mean of gradients with inconsistent signs. The server-client communication protocol of our proposed methods are further demonstrated in Figure \ref{fedilc} and \ref{sequence}.

\textbf{Problem Formulation:} To begin with, we define our problem in the following formulations. We assume there are $\mathcal{E}$ accessible clients in the federated learning settings. Each client $e \in \mathcal{E}$ has private training dataset $\{ \mathcal{D}_e \}_{e \in \mathcal{E}}$ sampled from $\mathcal{E}$ distinct distributions. Let $w$ be the parameters of the global model shared across clients. Our goal is to learn $w$ such that the model captures invariant features across different federated clients and generalises well to unseen domains.

\subsection{Weighted Geometric Mean}

Parascandolo introduced geometric averaging of Hessians in his paper \cite{parascandolo2021ilc}. Although the element-wise geometric mean of gradients successfully promotes consistency in convex quadratic cases, it is still circumscribed with several quandaries. One of the significant encumbrances is that \textbf{geometric mean is only defined when all the signs are consistent}, and it is barely feasible to calculate geometric mean with inconsistent signs, which pervasively exists in non-convex settings. Based on Parascandolo's code implementation, they take \textbf{the absolute values} of all the gradients to calculate the geometric mean. Though applied with a list of binary masks, the gradients calculated by geometric mean with absolute values might still introduce bias during the training process. Therefore, this method can be further improvable.

In this paper, we introduce \textbf{weighted geometric mean}, which remedies the conventional geometric mean method to compute when the signs are inconsistent. 

Let our global model be parametrised by $w \in \mathbf{R}^n$. Let $F_e$ be the loss obtained for client $e$, with its corresponding gradient $\nabla F_e$. 

Let $\left[ \nabla  F_e  \right]_k$ denote the k-th coordinate of $\nabla  F_e$,  $k \in \{0,.., n - 1 \}$ ($\left[ \nabla  F_e  \right]_k = \frac{\partial F_e}{\partial \theta_k} $).

For simplicity, let us denote $\left[ \nabla  F_e  \right]_k$ with $G_e$. And let $G  = \{ G_e , e \in \mathcal{E} \}$. We want to define the weighted geometric mean for $G$, by considering separately its positive and negative coordinates.

Let $\mathcal{E}^+ = \{ e \in \mathcal{E}, G_e \geq 0  \}$ , and $\mathcal{E}^- = \{ e \in \mathcal{E}, G_e \leq 0  \}$.

The arithmetic mean of $G$ satisfies:

\[ arithmean(G) =  \frac{\mathcal{E}^+}{\mathcal{E}}  \frac{\sum_{e \in \mathcal{E^+}} \left |   G_e   \right|}{\mathcal{E}^+}   -   \frac{\mathcal{E}^-}{\mathcal{E}}  \frac{\sum_{e \in \mathcal{E^-}} \left |   G_e   \right|}{\mathcal{E}^-}  \]

Similarly, We define the weighted geometric mean of $G$ as :

\[ weightedgeo(G) =  \frac{\mathcal{E}^+}{\mathcal{E}} \left( \prod_{e \in \mathcal{E^+}} \left |   G_e   \right| \right)^{\frac{1}{\mathcal{E}^+}} - \frac{\mathcal{E}^-}{\mathcal{E}} \left( \prod_{e \in \mathcal{E^-}} \left|   G_e  \right| \right)^{\frac{1}{\mathcal{E}^-}}  \]

This definition allows us to circumvent the problem of sign inconsistency for geometric mean.

Therefore, the weighted geometric mean of the gradient $\left[ \nabla^{\wedge}F \right]_k$ for the k-th parameter of the global model is calculated as follows:

\[ \left[ \nabla^{\wedge}F \right]_k =  \frac{\mathcal{E}^+}{\mathcal{E}} \left( \prod_{e \in \mathcal{E^+}} \left | \left[ \nabla  F_e  \right]_k \right| \right)^{\frac{1}{\mathcal{E}^+}} - \frac{\mathcal{E}^-}{\mathcal{E}} \left( \prod_{e \in \mathcal{E^-}} \left| \left[ \nabla F_e \right]_k \right| \right)^{\frac{1}{\mathcal{E}^-}}  \]

Then we update the global model in the federated networks based on the calculated geometric mean of gradients: $w  \leftarrow w - \eta \nabla^{\wedge} F$, where $\nabla^{\wedge}F = \{\nabla^{\wedge}F_k , k \in \{0,.., n-1\}  \}$. The implementation of weighted geometric mean implementation is further illustrated at 
\url{https://colab.research.google.com/drive/17y6ZuwiRE3iHVhFxaXz1b9eSZP7x25Om?usp=sharing}.

\subsection{Fishr+Inter-Geo: Fishr and Inter-silo Weighted Geometric Mean}

Rame proposed another gradient-based regularization method for domain generalization, \textbf{Fishr}, which matches the second-order information of gradient distributions - gradient covariance across different environments \cite{rame2021fishr}. 

However, since it is of paramount importance to know the number of environments beforehand for gradient covariance calculation, Rame hypothesized that \textbf{the number of domains is accessible} during training, which is barely tractable in a centralized way of training. However, we could utilize Fishr in the \textbf{federated learning} settings, where the number of federated clients $\mathcal{E}$ are antecedently accessible. 

To compute Fishr in a federated way, we first define the gradient covariance matrix, where $\nabla F_e^i$ denotes the first-order derivative for the $i$-th data example from client $e \in \mathcal{E}$, and $\overset{-}{\nabla F_e} = \frac{1}{n_e} \sum_{i=1}^{n_e} \nabla F_e^i$ denotes the gradient means of client $e$. The computed gradient covariance will be uploaded to server through RESTful APIs in each federated training round.

\[ C_e = \frac{1}{n_e} \sum_{i=1}^{n_e} \left( \nabla F_e^i - \overset{-}{\nabla F_e} \right) \left( \nabla F_e^i - \overset{-}{\nabla F_e} \right)^{\top} \]

Then we compute the Fishr loss on the centralized server, where the mean covariance matrix is $\overset{-}{C} = \frac{1}{\mathcal{E}} \sum_{e \in \mathcal{E}} C_e$:

\[ L_{Fishr} = \frac{1}{\mathcal{E}} \sum_{e \in \mathcal{E}} \Vert C_e - \overset{-}{C} \rVert^2 \]

Ultimately, we further ameliorate our Federated Fishr method with weighted geometric mean $\nabla^{\wedge} F_t$ in lieu of arithmetic averaging of gradients $\nabla^+ F_t$ uploaded by clients. Specifically, we utilise both inter-silo gradient covariance and domain-level weighted geometric mean to update the global model with the gradient descent algorithm. Given that both the Fishr and geometric averaging of Hessians are potent in capturing domain-level invariance, our proposed method, as proved in our experiments, further enforces the consistencies among federated clients. The pseudo-code is demonstrated in Algorithm \ref{alg:inter}.

\[ w_{t+1} = w_t - \eta \left( \nabla^{\wedge} F_t + \lambda \nabla L_{Fishr_t} \right) \]

\subsection{Fishr+Intra-Geo: Fishr and Intra-silo Weighted Geometric Mean}

Theoretically, the dataset within each client is more statistically homogeneous. In the real world cases, however, it is prevalent that \textbf{non-i.i.d. data may exist within one federated client}, which could potentially hamper the effective training of federated models. For example, the patients' clinical data, such as age or race, may significantly vary in different distributions, even within one hospital. To the best of our knowledge, this issue has not been solved by any previous federated learning research works.

Therefore, we further integrate \textbf{intra-silo weighted geometric mean} with inter-silo gradient covariance to mitigate this issue. Specifically, we calculate intra-silo weighted geometric mean of gradients by treating each data example $\left( x_e^i, y_e^i \right)$ as an independent environment in client $e$. Meanwhile, we compute gradient covariance in the same federated client. Later, each client collaboratively uploads the computed gradients and covariance information to the orchestration of the central server to update the global model's parameters. The pseudo-code is demonstrated in Algorithm \ref{alg:intra}.

\section{Theoretical Analysis}

Parascandolo's paper introduces the Inconsistency Score \cite{parascandolo2021ilc}. Suppose we have two clients $A$ and $B$ with distinct training distributions, then the Inconsistency Score $I\left(\theta^*\right)$ is defined as follows, where $\theta^*$ denotes the local minimum of the loss space $L$.

\[I\left(\theta^*\right) = max \left\{ \underset{\lvert L_A(\theta) - L_A(\theta^*) \rvert \leq \epsilon }{max} \lvert L_B \left( \theta \right) - L_A \left( \theta^* \right) \rvert, \underset{\lvert L_B(\theta) - L_B(\theta^*) \leq \epsilon }{max} \lvert L_A \left( \theta \right) - L_B \left( \theta^* \right) \rvert \right\}\]





Based on the proof in Parascandolo's paper \cite{parascandolo2021ilc}, the Inconsistency Score is lowest when both $A$ and $B$ have the same curvature in Hessians: $I\left(\theta^*\right) = \epsilon \cdot max \underset{i}{max} \left\{ \frac{\lambda_i^B}{\lambda_i^A}, \frac{\lambda_i^A}{\lambda_i^B} \right\} $ where $H_A$ and $H_B$ have the similar eigenvalues.

Our proposed methods, including both Fishr+Inter-Geo and Fishr+Intra-Geo, successfully reduces the Inconsistency Score $I\left(\theta^*\right)$, thus promotes the consistencies across different environments. The justifications are as follows:

\begin{itemize}
    \item Based on Parascandolo's paper, by employing the geometric mean $H^{\wedge}$ in lieu of the arithmetic mean $H^{+}$ of Hessians, we can \textbf{reduce the training speed} of convergence on directions where landscapes with different curvatures might lead to a high inconsistency in Hessians \cite{parascandolo2021ilc}. Therefore, the model is more likely to find the consistent minimizers, where their Hessians have similar eigenvalues $\lambda_i^A$ and $\lambda_i^B$, thus minimizing the Inconsistency Score $I\left(\theta^*\right)$.
    \item Based on Fishr paper, the gradient covariance $C$ is equivalent to the Fishr Information Matrix $\Tilde{F}$ at any first-order stationary point: $C {\propto} \Tilde{F}$. Further, many research works have proved that the Fishr Information Matrix $\Tilde{F}$ and Hessians $H$ share the same structure and are similar up to a scalar factor: $\Tilde{F} {\propto} H$ \cite{frantar2021efficient,liu2021group,dangel2021curvature}. Therefore, \textbf{the gradient covariance $C$ is closely related to Hessians $H$ }: $C {\propto} H$. Therefore, employing Fishr loss can further enforce Hessians $H$ to be similar, thus minimizing the Inconsistency Score $I\left(\theta^*\right)$.
\end{itemize}

\section{Experiments}

We conduct extensive experiments of FedILC on the Color-MNIST dataset \cite{arjovsky2020irm}, Rotated-CIFAR10 dataset, and the real-world eICU dataset. Further, we compare the performance of six different methodologies, including the conventional baseline algorithm FedSGD, the state-of-the-art baseline algorithm FedCurv, two baseline algorithms developed by ourselves, and our nascently proposed algorithms Fishr+Inter-Geo and Fishr+Intra-Geo:

\textbf{Baseline methods:}
\begin{itemize}
  \item FedSGD: the baseline federated learning algorithm which performs the arithmetic mean of the uploaded clients' gradients.
  \item FedCurv: the state-of-the-art baseline federated learning algorithm which integrates with vanilla Fishr regularization.
  \item Geometric: the federated learning algorithm which performs the weighted geometric mean of the uploaded clients' gradients.
  \item Fishr+Intra-Arith: Fishr and intra-silo arithmetic mean algorithm for comparison purpose.
\end{itemize}

\textbf{Proposed methods:}
\begin{itemize}
  \item Fishr+Inter-Geo: Fishr and inter-silo weighted geometric mean algorithm.
  \item Fishr+Intra-Geo: Fishr and intra-silo weighted geometric mean algorithm.
\end{itemize}

Notably, as suggested by Rame in his paper, it is sufficient to calculate Fishr with the gradients in the final linear classifier, which essentially reduces the memory overhead \cite{rame2021fishr}. Furthermore, research has manifested that low-level layers are paramount to domain-dependent peculiarities, and the final layers of a neural network are most prone to overfit to the client distribution \cite{luo2021heterogeneity}. Rame also mentioned that we could further scale down the computational cost by applying the gradient variance, the diagonal of the covariance matrix $Diag(C)$, because the empirical evidence has unveiled that the diagonal of Hessians $Diag(H)$ are prominent at the end of training \cite{sue1988improving,adolphs2020ellipsoidal,singh2020woodfisher}. Therefore, in our implementation of Fishr, we only consider the gradient variance in lieu of covariance in the final linear classifier of each client's model. Moreover, regarding the Fishr and inter-silo weighted geometric mean algorithm (Fishr+Inter-Geo), computing per-sample weighted geometric mean may also become computationally expensive, especially when the federated client has a copious amount of data. Therefore, we purvey a scalable solution by calculating the weighted geometric mean of mini-batch gradients within each client, which largely mitigates the computational overhead.

\subsection{Environment Setup}

\subsubsection{Color-MNIST}

In the Color-MNIST dataset \cite{arjovsky2020irm}, the color of digits is spuriously pertinent to the labels. We assume five clients in the federated networks with different feature heterogeneity, whose color of digits is randomly flipped with a probability of 15\%, 30\%, 45\%, 60\%, and 75\%, respectively. We also create an OOD test set with a color flipping probability of 90\% and a label flipping probability of 15\% to evaluate the generalizability of the global model. The goal is to solve a binary classification problem and predict whether a given digit is below or above 5. Therefore, we create a 3-layer MLP model with 390, 390, and 1 neuron in the corresponding layer. We also adopt the Adam optimizer with a fixed learning rate of 0.0003 and weight decay of 0.01. Critically, we choose the coefficient $\lambda = 15.0$ to control the Fishr regularization strength.

\subsubsection{Rotated-CIFAR10}

\begin{wraptable}{r}{8cm}
\caption{Experiment setup of Rotated-CIFAR10 involves three clients with distinct rotation degrees. Each client also contains three sub-environments to simulate intra-silo feature heterogeneity.}
\label{table:cifar}
\begin{tabular}{@{}llll@{}}
\toprule
         & Sub Env 1 & Sub Env 2 & Sub Env 3 \\ \midrule
Client 1 & 10°   & 25°   & 40°   \\
Client 2 & 60°   & 75°   & 90°   \\
Client 3 & -10°  & -40°  & -90°  \\ \bottomrule
\end{tabular}
\end{wraptable}

Based on the conventional CIFAR10 dataset \cite{krizhevsky2009cifar}, we also develop our own synthetic dataset - Rotated-CIFAR10 as our benchmark dataset. We rotate CIFAR10 images with different angles to simulate the non-i.i.d. data in the federated networks. In our federated learning settings, there are three clients with distinct data distributions. Further, we also establish three sub-environments in each client to simulate intra-silo feature heterogeneity. The experiment setup is demonstrated in Table \ref{table:cifar}. Moreover, the OOD test set is configured with a rotation degree from -90° to 90° to evaluate the performance of the global model. Our goal is to correctly classify the 32x32 RGB colored images with 10 classes. Therefore, we adopt a 3-layer CNN architecture as our classification model. We also use the Adam optimizer with a fixed learning rate of 0.0001 and a weight decay of 0.001. Particularly, we select $\lambda = 1.0$ as the Fishr hyperparameter.

\subsubsection{eICU}

The eICU Collaborative Research Database is an extensive real-world medical dataset created by Philips Healthcare in partnership with the MIT Laboratory for Computational Physiology \cite{pollard2018eicu}. The dataset is spatially distributed in 58 hospitals, with 30,760 patients' information about the medicines taken and the mortality of each patient. Furthermore, the dataset is imbalanced, with 30.5\% of patients recorded as dead in our dataset while others recorded as alive. We also select 20 among the hospitals with the most data to participate in the training process, whilst the rest are the OOD test set to evaluate the global model. Moreover, we create a training set in each training hospital with 70\% of its data. Meanwhile, the rest of 30\% of the data becomes the validation set. Our objective is to predict a patient's mortality based on the 1399 kinds of medicines taken in as input features. Therefore, we construct a 4-layer MLP with 1024, 1024, 512, and 1 neuron to perform a binary classification task. We also use the Adam optimizer with a fixed learning rate of 0.0002 and a weight decay of 0.001. Besides, the Fishr regularizer $\lambda$ we choose is 0.1 during the training process.

\subsection{Result and Analysis}

\begin{wrapfigure}{r}{5cm}
  \centering
  {\setlength{\fboxsep}{0pt}%
  \setlength{\fboxrule}{0pt}%
  \fbox{\includegraphics[width=4cm, height=3cm]{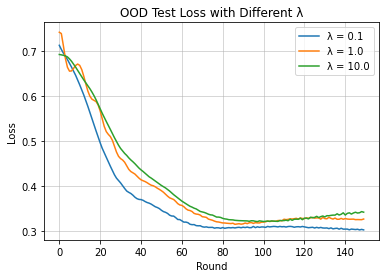}}
}%
  \caption{The OOD test loss with different Fishr regularizer $\lambda$ on the eICU dataset.}
 \label{lambda_compare}
\end{wrapfigure}

\begin{figure}[!htb]
  \centering
  {\setlength{\fboxsep}{0pt}%
  \setlength{\fboxrule}{0pt}%
  \fbox{\includegraphics[width=13.5cm, height=3.375cm]{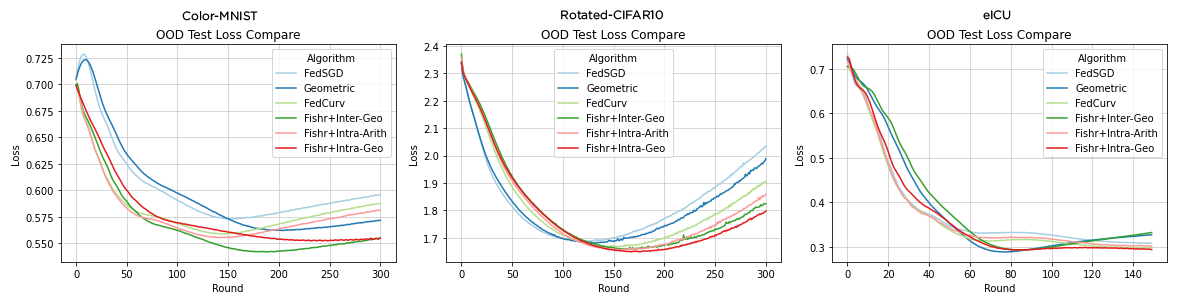}}
}%
  \caption{The OOD test loss comparison among different algorithms on the Color-MNIST, Rotated-CIFAR10, and eICU dataset.}
 \label{experiment_loss_compare_figure}
\end{figure}

We run each algorithm for five different seeds, then average their loss, the AUCROC, AUCPR, as well as the accuracy on the OOD test set when the OOD test loss is lowest, which are demonstrated in Table \ref{experiment_loss_compare_table}, Table \ref{experiment_roc_pr_compare_table} Table \ref{experiment_accuracy_compare_table}, and Figure \ref{experiment_loss_compare_figure}. Further, we also conduct the inter-silo and intra-silo accuracy comparison among each sub environment on the Rotated-CIFAR10 dataset when the OOD test loss is lowest \ref{experiment_inter_intra_accuracy_table}. Specifically, we also experiment with the model's performance regarding different Fishr regularizer $\lambda$ on the eICU dataset \ref{lambda_compare}. Moreover, we also plot the training loss, validation loss, and OOD test loss among different algorithms, which are demonstrated in Figure \ref{mnist_loss}, Figure \ref{cifar_loss}, and Figure \ref{icu_loss} in the Appendix section. From the figures, our proposed methods, Fishr+Inter-Geo and Fishr+Intra-Geo, significantly outperform other methods in domain adaptation, including the baseline algorithm FedSGD and the state-of-the-art algorithm FedCurv.

Last but not least, we also measure the fairness of our proposed algorithm by evaluating (1) the variance of the accuracy distribution and (2) the KL-divergence between the normalized accuracy and the uniform distribution, which can be directly translated to the entropy of accuracy defined in the paper \cite{li2020fair}. The result of our experiments is demonstrated in Table \ref{experiment_variance_compare_table}.

\section{Conclusion}

In conclusion, to mitigate the domain shift problem engendered by the non-i.i.d. data in the federated learning settings, we propose the Federated Invariant Learning Consistency (FedILC), including both Fishr+Inter-Geo and Fishr+Intra-Geo methodologies. Furthermore, we also introduce a novel weighted geometric mean strategy to calculate the geometric mean of gradients with inconsistent signs. The benchmark and real-world dataset experiments demonstrate that our proposed algorithm outperforms both the conventional and the state-of-the-art federated learning algorithms in capturing both inter-silo and intra-silo consistencies in federated networks. 

\clearpage

\begin{table}[!htb]
\caption{The OOD test loss comparison among different algorithms on the Color-MNIST, Rotated-CIFAR10, and eICU dataset. }
\label{experiment_loss_compare_table}
\centering
\begin{tabular}{@{}llll@{}}
\toprule
                         & Color-MNIST & Rotated-CIFAR10 & eICU                 \\ \toprule
FedSGD                   & 0.566±0.009 & 1.681±0.010     & 0.302±0.017          \\ \midrule
Geometric                & 0.562±0.006 & 1.681±0.006     & \textbf{0.289±0.009} \\ \midrule
FedCurv                  & 0.559±0.003 & 1.666±0.024     & 0.298±0.009          \\ \midrule
\textbf{Fishr+Inter-Geo} & \textbf{0.542±0.006} & 1.658±0.004 & 0.293±0.009          \\ \midrule
Fishr+Intra-Arith        & 0.555±0.004 & 1.655±0.007     & 0.302±0.009          \\ \midrule
\textbf{Fishr+Intra-Geo} & 0.552±0.005 & \textbf{1.648±0.006} & \textbf{0.289±0.008} \\ \bottomrule
\end{tabular}
\end{table}

\begin{table}[!htb]
\caption{The variance of the accuracy distribution and KL-divergence between the normalized accuracy vector and the uniform distribution (which can be directly translated to the entropy of accuracy) on the Rotated-CIFAR10 dataset when the OOD test loss is lowest.}
\label{experiment_variance_compare_table}
\centering
\begin{tabular}{@{}lll@{}}
\toprule
                         & Variance (x1000)     & Entropy (x10)         \\ \toprule
FedSGD                   & 0.809±0.251          & 21.948±0.007          \\ \midrule
FedCurv                  & \textbf{0.606±0.093} & 21.955±0.003          \\ \midrule
\textbf{Fishr+Inter-Geo} & 0.687±0.234          & 21.953±0.007          \\ \midrule
\textbf{Fishr+Intra-Geo} & 0.611±0.120          & \textbf{21.956±0.003} \\ \bottomrule
\end{tabular}
\end{table}

\begin{table}[!htb]
\caption{The AUCROC and AUCPR comparison among different algorithms on the Color-MNIST, Rotated-CIFAR10, and eICU dataset when the OOD test loss is lowest.}
\label{experiment_roc_pr_compare_table}
\begin{subtable}{0.55\linewidth}
\caption{AUCROC}
\begin{tabular}{lll}
\hline
                         & Color-MNIST          & eICU                 \\ \toprule
FedSGD                   & 0.780±0.001          & 0.556±0.004          \\ \midrule
Geometric                & 0.781±0.003          & 0.575±0.004          \\ \midrule
FedCurv                  & 0.780±0.003          & 0.556±0.005          \\ \midrule
\textbf{Fishr+Inter-Geo} & \textbf{0.789±0.002} & \textbf{0.577±0.005} \\ \midrule
Fishr+Intra-Arith        & 0.782±0.005          & 0.553±0.002          \\ \midrule
\textbf{Fishr+Intra-Geo} & 0.778±0.002          & 0.567±0.004          \\ \bottomrule
\end{tabular}
\end{subtable}%
\begin{subtable}{0.55\linewidth}
\caption{AUCPR}
\begin{tabular}{lll}
\hline
                         & Color-MNIST          & eICU                 \\ \toprule
FedSGD                   & 0.837±0.002          & 0.209±0.010          \\ \midrule
Geometric                & 0.838±0.004          & 0.214±0.008          \\ \midrule
FedCurv                  & 0.838±0.004          & 0.217±0.018          \\ \midrule
\textbf{Fishr+Inter-Geo} & \textbf{0.845±0.002} & 0.218±0.009          \\ \midrule
Fishr+Intra-Arith        & 0.839±0.004          & 0.216±0.016          \\ \midrule
\textbf{Fishr+Intra-Geo} & 0.836±0.001          & \textbf{0.221±0.013} \\ \bottomrule
\end{tabular}
\end{subtable}
\end{table}

\clearpage
\bibliographystyle{bib_style}
\bibliography{federated_ilc.bib}

\clearpage
\section{Appendix}

\begin{table}[htb]
\begin{tabular}{@{}ll@{}}
\toprule
Notation                                                            & Description                                                                      \\ \midrule
$\mathcal{E}$                                                       & Total number of clients in the federated networks                                \\
$e$                                                                 & An individual federated client in the federated networks                         \\
$\mathcal{D}$                                                       & Local dataset of client                                                          \\
$w$                                                                 & Global weights of server                                                      \\
$\left[ \mathcal{E}^+ \right]_i$ / $\left[ \mathcal{E}^- \right]_i$ & The number of clients with positive / negative gradients for $i$-th data example \\
$\nabla^{\wedge} F$                                                 & Weighted geometric mean of gradients                                             \\
$\nabla^+ F$                                                        & Arithmetic mean of gradients                                                     \\
$\eta$                                                              & Learning rate                                                                    \\
$\lambda$                                                           & Hyper-parameter to control the Fishr regularization strength                     \\
$C$                                                                 & Gradient covariance matrix                                                       \\ \bottomrule
\end{tabular}
\end{table}

\begin{table}[!htb]
\caption{The inter-silo and intra-silo accuracy comparison on the Rotated-CIFAR10 dataset when the OOD test loss is lowest. }
\label{experiment_inter_intra_accuracy_table}
\centering
\begin{tabular}{@{}lllll@{}}
\toprule
                   & FedSGD      & FedCurv     & \textbf{Fishr+Inter-Geo} & \textbf{Fishr+Intra-Geo} \\ \toprule
Client 1 Sub Env 1 & 0.443±0.010 & 0.435±0.013 & 0.446±0.020          & \textbf{0.456±0.017} \\ \midrule
Client 1 Sub Env 2 & 0.434±0.017 & 0.448±0.008 & 0.438±0.009          & \textbf{0.451±0.011} \\ \midrule
Client 1 Sub Env 3 & 0.428±0.008 & 0.424±0.006 & 0.442±0.013          & \textbf{0.444±0.009} \\ \midrule
Client 2 Sub Env 1 & 0.420±0.020 & 0.425±0.004 & 0.435±0.004          & \textbf{0.441±0.008} \\ \midrule
Client 2 Sub Env 2 & 0.440±0.009 & 0.437±0.011 & \textbf{0.443±0.009} & 0.438±0.008          \\ \midrule
Client 2 Sub Env 3 & 0.411±0.007 & 0.409±0.005 & 0.404±0.008          & \textbf{0.419±0.011} \\ \midrule
Client 3 Sub Env 1 & 0.413±0.014 & 0.431±0.011 & 0.427±0.017          & \textbf{0.443±0.004} \\ \midrule
Client 3 Sub Env 2 & 0.367±0.010 & 0.379±0.014 & \textbf{0.392±0.012} & 0.388±0.012          \\ \midrule
Client 3 Sub Env 3 & 0.371±0.010 & 0.380±0.010 & 0.381±0.016          & \textbf{0.396±0.010} \\ \bottomrule
\end{tabular}
\end{table}

\begin{table}[!htb]
\caption{The accuracy comparison among different algorithms on the Color-MNIST, Rotated-CIFAR10, and eICU dataset when the OOD test loss is lowest. Since the eICU dataset is highly imbalanced, the accuracy on the eICU dataset is not fully informative.}
\label{experiment_accuracy_compare_table}
\centering
\begin{tabular}{@{}llll@{}}
\toprule
                         & Color-MNIST          & Rotated-CIFAR10      & eICU                 \\ \toprule
FedSGD                   & 0.781±0.001          & 0.404±0.004          & 0.919±0.008          \\ \midrule
Geometric                & 0.781±0.003          & 0.405±0.003          & 0.900±0.005          \\ \midrule
FedCurv                  & 0.780±0.004          & 0.408±0.010          & \textbf{0.923±0.005} \\ \midrule
\textbf{Fishr+Inter-Geo} & \textbf{0.790±0.002} & 0.414±0.003          & 0.907±0.003          \\ \midrule
Fishr+Intra-Arith        & 0.783±0.005          & 0.414±0.003          & \textbf{0.923±0.006} \\ \midrule
\textbf{Fishr+Intra-Geo} & 0.779±0.002          & \textbf{0.419±0.004} & 0.917±0.006          \\ \bottomrule
\end{tabular}
\end{table}

\clearpage
\begin{algorithm*}

\caption{Fishr+Inter-Geo: Fishr and Inter-silo Weighted Geometric Mean.}\label{alg:inter}

\begin{algorithmic}

\Function{ServerExecutes}{\null}

  \State initialize $w = w_0 \in \mathbf{R}^{n}$
  \For{each round $t$ = 0, 1, 2, ...}
    
    \For{each client $e \in \mathcal{E}$}
        \State $\nabla F_{e}, C_{e} \gets$ ClientUpdates($e$, $w$)
    \EndFor
    
    \State $G_k \gets \{ [\nabla F_e]_k, e \in \mathcal{E} \} , k \in[0..n-1]$
    
    \State $\left[ \nabla^{\wedge}F \right]_k \gets weightedgeo(G_k), k \in [0..n-1]$ \Comment{Weighted geometric mean}
    
    \State $L_{Fishr} \gets \frac{1}{\mathcal{E}} \sum_{e \in \mathcal{E}} \Vert C_{e} - \overset{-}{C} \rVert^2 $ \Comment{Fishr loss}
    
    \State $w \gets w - \eta \left( \nabla^{\wedge} F + \lambda \nabla L_{Fishr} \right)$
  \EndFor
\EndFunction
\\
\Function{ClientUpdates}{$e$, $w$}

  \State $\nabla F_e \gets$ backwards($F_e$, $w$)
  \State $C_e \gets \frac{1}{n_e} \sum_{i=1}^{n_e} \left( \nabla F_e^i - \overset{-}{\nabla F_e} \right) \left( \nabla F_e^i - \overset{-}{\nabla F_e} \right)^{\top}$ \Comment{Gradient covariance}
  \State \Return $\nabla F_e$, $C_e$
\EndFunction

\end{algorithmic}
\end{algorithm*}

\begin{algorithm*}

\caption{Fishr+Intra-Geo: Fishr and Intra-silo Weighted Geometric Mean.}\label{alg:intra}

\begin{algorithmic}

\Function{ServerExecutes}{\null}

  \State initialize $w_0$
  \For{each round $t$ = 0, 1, 2, ...}
    
    \For{each client $e \in \mathcal{E}$}
        \State $\nabla^{\wedge} F_e, C_{e} \gets$ ClientUpdates($e$, $w$)
    \EndFor
    
    \State $\nabla F \gets arithmean(\nabla^{\wedge} F_e, e \in \mathcal{E})$
    
    \State $L_{Fishr} \gets \frac{1}{\mathcal{E}} \sum_{e \in \mathcal{E}} \Vert C_{e} - \overset{-}{C} \rVert^2 $ \Comment{Fishr loss}
    
    \State $w \gets w - \eta \left( \nabla F + \lambda \nabla L_{Fishr} \right)$
  \EndFor
\EndFunction
\\
\Function{ClientUpdates}{$e$, $w$}

  \For{each data example $\left( x_e^i, y_e^i \right) \in \mathcal{D}_e$}
    \State $\nabla F_e^i \gets$ backwards($F_e^i$, $w$)
  \EndFor

  \State $G_{e,k} \gets \{ [\nabla F_e^i]_k, i \in n_e \}, k \in [0..n-1]$

  \State $[\nabla^{\wedge}F_e]_k \gets weightedgeo(G_{e,k}) , k \in [0..n-1]$ \Comment{Weighted geometric mean}
  
  \State $C_e \gets \frac{1}{n_e} \sum_{i=1}^{n_e} \left( \nabla F_e^i - \overset{-}{\nabla F_e} \right) \left( \nabla F_e^i - \overset{-}{\nabla F_e} \right)^{\top}$ \Comment{Gradient covariance}
  \State \Return $\nabla^{\wedge} F_e$, $C_e$
\EndFunction

\end{algorithmic}
\end{algorithm*}

\clearpage

\clearpage
\begin{figure}[!htb]
  \centering
  {\setlength{\fboxsep}{0pt}%
  \setlength{\fboxrule}{0pt}%
  \fbox{\includegraphics[width=11.5cm, height=5.5cm]{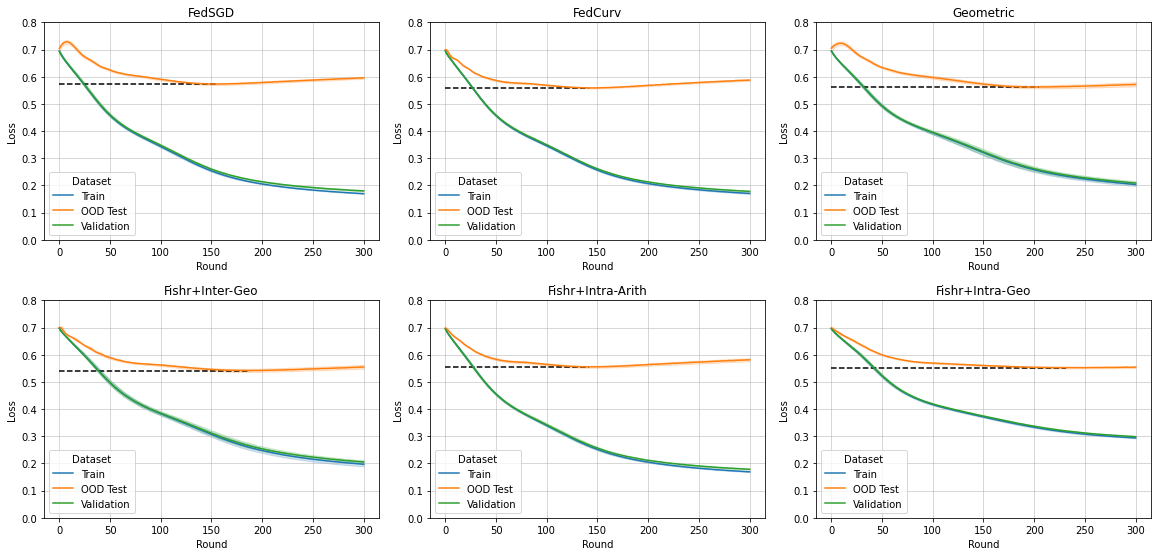}}
}%
  \caption{The training loss, validation loss, and OOD test loss comparison among different algorithms on the Color-MNIST dataset.}
 \label{mnist_loss}
\end{figure}

\begin{figure}[!htb]
  \centering
  {\setlength{\fboxsep}{0pt}%
  \setlength{\fboxrule}{0pt}%
  \fbox{\includegraphics[width=11.5cm, height=5.5cm]{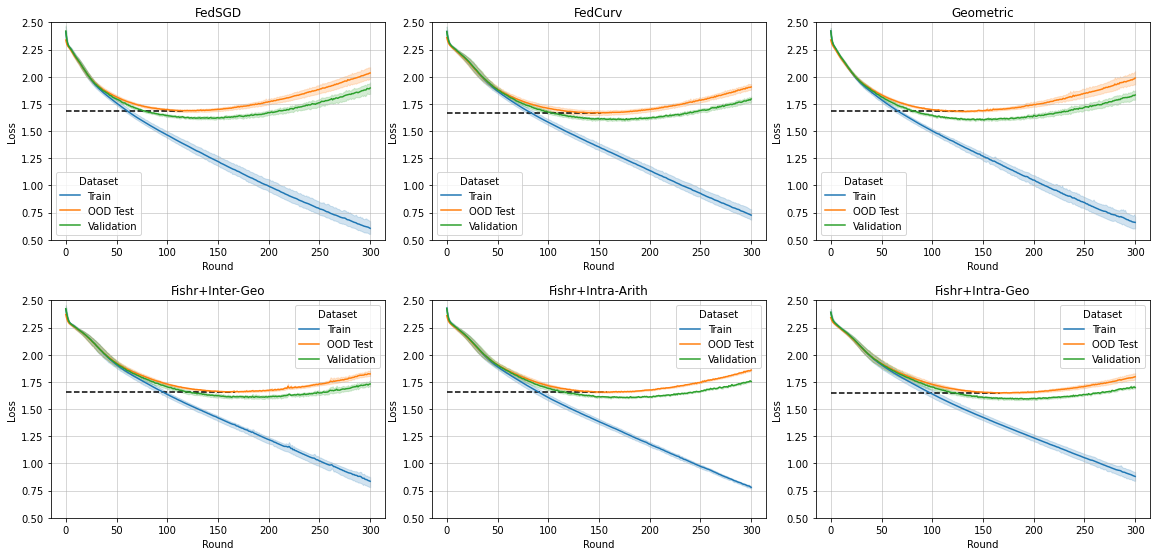}}
}%
  \caption{The training loss, validation loss, and OOD test loss comparison among different algorithms on the Rotated-CIFAR10 dataset.}
 \label{cifar_loss}
\end{figure}

\begin{figure}[!htb]
  \centering
  {\setlength{\fboxsep}{0pt}%
  \setlength{\fboxrule}{0pt}%
  \fbox{\includegraphics[width=11.5cm, height=5.5cm]{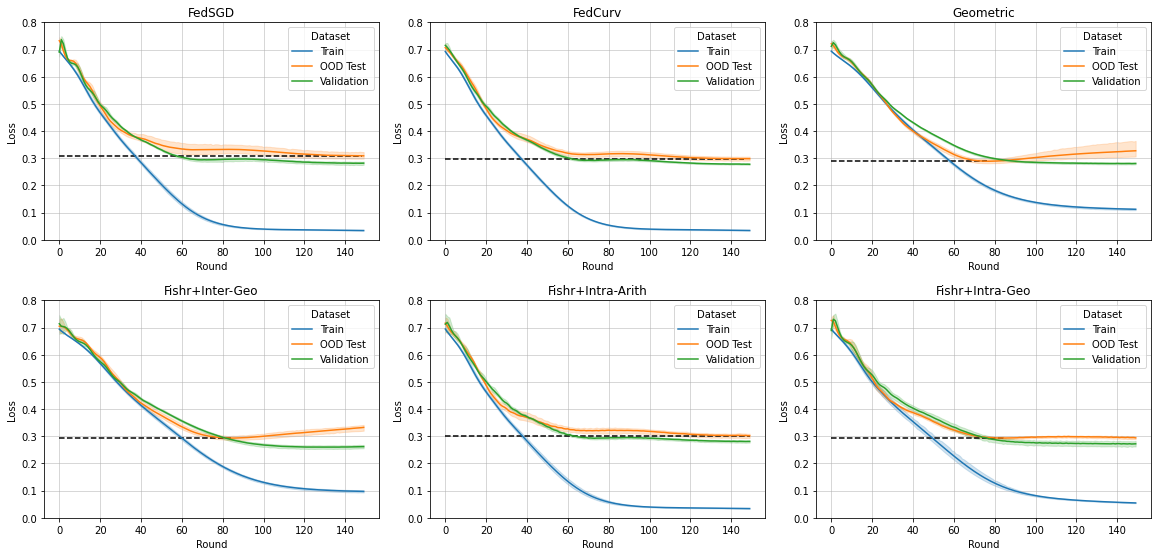}}
}%
  \caption{The training loss, validation loss, and OOD test loss comparison among different algorithms on the eICU dataset.}
 \label{icu_loss}
\end{figure}

\end{document}